\documentclass[sigconf]{acmart}
\usepackage{enumerate} 
\usepackage{multirow}
\newcommand{\etal}{\textit{et al.}}
\newcommand{\eat}[1]{}
\usepackage{color}
\usepackage{url}
\usepackage[linesnumbered,ruled,vlined]{algorithm2e}
\SetKwInput{KwInput}{Inputs}                
\SetKwInput{KwOutput}{Outputs}              

\AtBeginDocument{%
\providecommand\BibTeX{{%
		\normalfont B\kern-0.5em{\scshape i\kern-0.25em b}\kern-0.8em\TeX}}}

\setcopyright{acmcopyright}
\copyrightyear{2018}
\acmYear{2018}
\acmDOI{XXXXXXX.XXXXXXX}

\acmConference[Conference acronym 'XX]{Make sure to enter the correct
conference title from your rights confirmation emai}{June 03--05,
2018}{Woodstock, NY}
\acmPrice{15.00}
\acmISBN{978-1-4503-XXXX-X/18/06}

\begin{document}

\title{Skeleton-based Action Recognition via Adaptive \\ Cross-Form  Learning}
\fancyhead{}

\author{Xuanhan Wang}
\affiliation{%
	\institution{\textbf{}}
	\country{\textbf{}}
}
\author{Yan Dai}
\affiliation{%
	\institution{\textbf{}}
	\country{\textbf{}}
}
\author{Lianli Gao}
\affiliation{%
	\institution{\textbf{}}
	\country{\textbf{}}
}
\author{Jingkuan Song}
\affiliation{%
	\institution{\textbf{}}
	\country{\textbf{}}
}

\begin{abstract}
	Skeleton-based action recognition aims to project skeleton sequences to action categories, where skeleton sequences are derived from multiple forms of pre-detected points. Compared with earlier methods that focus on exploring single-form skeletons via Graph Convolutional Networks (GCNs), existing methods tend to improve GCNs by leveraging multi-form skeletons due to their complementary cues. However, these methods (either adapting structure of GCNs or model ensemble) require the co-existence of all forms of skeletons during both training and inference stages, while a typical situation in real life is the existence of only partial forms for inference. To tackle this issue, we present Adaptive Cross-Form Learning (ACFL), which empowers well-designed GCNs to generate complementary representation from single-form skeletons without changing model capacity. Specifically, each GCN model in ACFL not only learns action representation from the single-form skeletons, but also adaptively mimics useful representations derived from other forms of skeletons. In this way, each GCN can learn how to strengthen what has been learned, thus exploiting model potential and facilitating action recognition as well. Extensive experiments conducted on three challenging benchmarks, i.e., NTU-RGB+D 120, NTU-RGB+D 60 and UAV-Human, demonstrate the effectiveness and generalizability of the proposed method. Specifically, the ACFL significantly improves various GCN models (i.e., CTR-GCN, MS-G3D, and Shift-GCN), achieving a new record for skeleton-based action recognition. 

\end{abstract}



\maketitle

\section{Introduction}
\label{sec:intro}
Human action recognition is a fundamental yet challenging task as it supports various downstream applications range from intelligent video surveillance~\cite{surveillance_mm} to human-computer interaction~\cite{Human_Interaction,human_computer_duetmusic_mm}. To identify actions accurately, various modalities of video have been explored, such as RGB images~\cite{rgb_eccv}, optical flows~\cite{optical_flows,GAO2020107477}, audio waves~\cite{audio_waves} and skeleton sequences~\cite{ST-GCN_AAAI}. In particular, the skeleton based modality has gained significance over other modalities due to its robustness against complicated backgrounds and its suitability for edge devices. Therefore, numerous skeleton-based action recognition methods have been proposed~\cite{LSTM_skeleton,PoseC3D,CTR_ICCV2021,ST-GCN_AAAI}, which aim to project skeleton sequences to action categories. 

\begin{figure}[t]
	\setlength{\belowcaptionskip}{-0.3cm}%
	\begin{center}
		\includegraphics[width=1.0\linewidth]{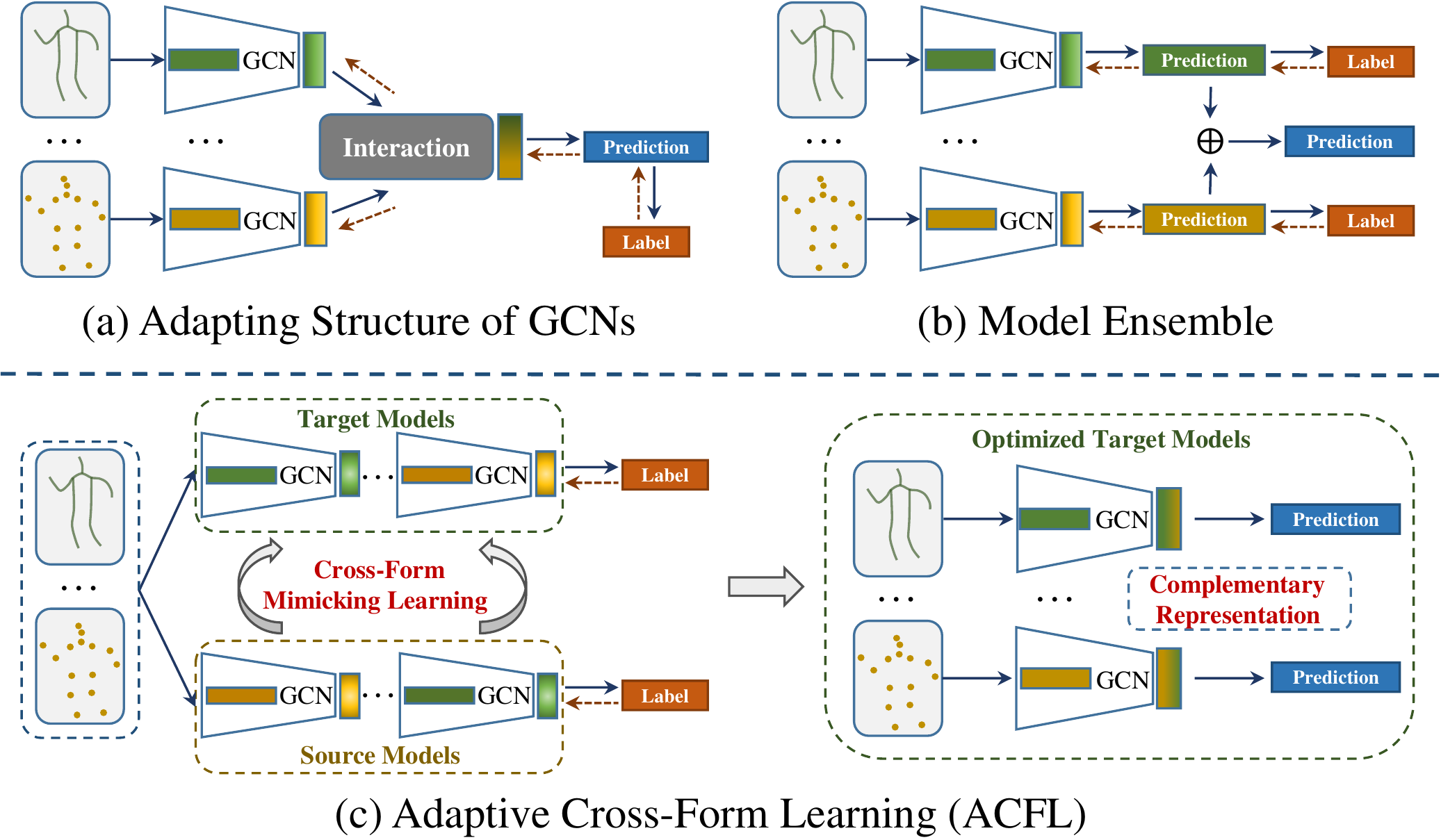}
	\end{center}
	\caption{Existing methods require the co-existence of multiple forms of skeletons during both training and inference stages: (a) Adapting structure of GCNs, where an interaction module embedded in GCN is used to fuse various representations derived from different single-form models; (b) Model ensemble, where it fuses various single-form skeleton representations by late fusion. Different from them, our goal is to empower a well-trained GCN model to generate complementary from partial forms of skeletons in the inference stage. To achieve this, our method (c) forces a GCN model not only to learn single-form skeleton based representation, but also to adaptively mimic useful representations derived from different skeleton forms during training. 
	}
	\label{fig:teaser}
\end{figure}

In general, skeleton sequences can be cast as spatial-temporal graphs for action recognition, as they are usually derived from multiple forms of pre-detected points (e.g., joint, bone, or both of them) via keypoint detection methods \cite{pose_SunXLWang2019,pose:stip}. Specifically, joint-based skeletons record the temporal variation of coordinates for all anatomical human keypoints, while kinetic keypoint chain is often characterized by bone-based skeletons. To model the spatial-temporal graphs, state-of-the-art methods focus on the elaborate design of Graph Convolutional Networks (GCNs)~\cite{SGN,Dynamic_GCN,ST-GCN_AAAI,Structural_gcn,MS_G3D,CTR_ICCV2021,shift_GCN,mix_skelton_mm}, aiming to learn action representation from single-form skeletons. However, only learning from single-form skeletons will lead to an unreliable result and limits the potential of GCNs, since identifying complex actions relies on complementary cues derived from multiple forms of skeletons. For instance, when identifying an action of \textit{``play magic cube''}, focusing on relative changes between fingers that are characterized by $bone$ based skeletons, would likely make a correct recognition. However, recognizing the action of \textit{``wear a shoe''}, prefers locations of hands as well as feet, which are characterized by $joint$ based skeletons. To address this, several works~\cite{2s-gcn,ResGCN,DGNN,joint2bone} attempt to utilize multiple forms of skeletons with complementary cues to obtain a reliable result. As illustrated in Fig.~\ref{fig:teaser}, these methods can be roughly divided into two directions: (1) Adapting structure of GCNs (Fig.~\ref{fig:teaser}(a)); and (2) Model ensemble (Fig.~\ref{fig:teaser}(b)). In the first direction, feature-level interaction modules are embedded in GCNs for generating complementary representations. As for the second direction, complementary representations are often obtained by a late fusing operation. Despite progress, these methods require the co-existence of all forms of skeletons during both training and inference stages, while a typical situation in real life is the existence of only partial forms for inference. 

\begin{figure}[t]
	
	\setlength{\belowcaptionskip}{-0.3cm}%
	\begin{center}
		\includegraphics[width=0.95\linewidth]{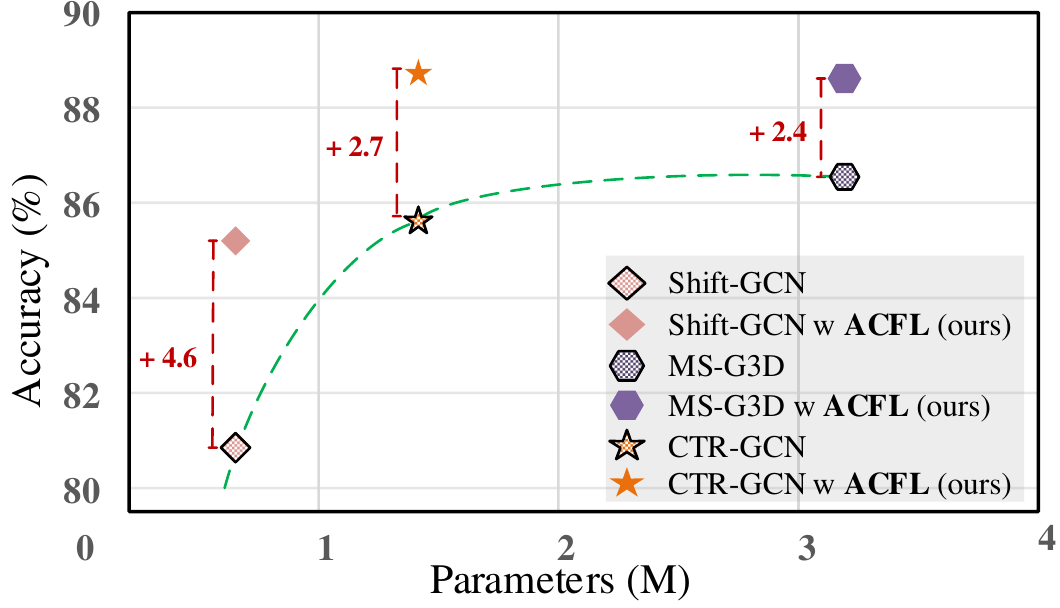}
	\end{center}
	\caption{Comparisons of different single-form GCN models on NTU120 (X-Sub) in terms of the recognition accuracy and the number of parameters. Enlarging model capacity via designing extra GCN modules brings relatively minor gains. For example, the complexity of MS-G3D is almost twice than that of CTR-GCN, but its recognition accuracy is improved by only 1.0\%. After applying proposed ACFL without changing model capacity, the performance of GCN models is significantly improved by ranging from 2.4\% to 4.6\%.}
	\label{fig:trend}
\end{figure}

Motivated by the above analysis, in this paper, we aim to answer one question: how to generate complementary representation from single-form skeletons for facilitating action recognition. To tackle this, we present Adaptive Cross-Form Learning (ACFL), a novel learning paradigm towards empowering well-designed GCNs to generate complementary representations from single-form skeletons without changing model capacity. Specifically, as shown in Fig.~\ref{fig:teaser}(c), each GCN model in ACFL not only learns action representation from the single-form skeletons, but also learns to adaptively mimic useful representations derived from other forms of skeletons. In this way, each GCN can smartly strengthen what has been learned and can produce complementary representation even from only one form of skeletons during inference, thus exploiting model potential and facilitating action recognition as shown in Fig.~\ref{fig:trend}. 

In summary, our work has three main contributions:
\begin{enumerate}[(1)]
	\item We propose a novel learning paradigm named adaptive cross-form learning (ACFL), which empowers a model to produce complementary action representations from partial skeleton forms aiming to relieve the requirement of the co-existence of all skeleton forms in the inference stage.
	\item The ACFL can be applied to optimize any GCN models for exploring their potential without changing model architecture, since it is model-agnostic and forces a model to adaptively mimic useful representations from various single-form models.
	\item Extensive experiments on three challenging benchmarks, namely, NTU-RGB+D 120, NTU-RGB+D 60 and UAV-Human, demonstrate the generalizability and effectiveness of our proposed ACFL. Without bells and whistles, it respectively achieves 2.5\%, 1.6\% and 2.2\% higher average accuracy (Acc) than CTR-GCN, MS-G3D and Shift-GCN on all benchmarks, achieving a new record for skeleton-based action recognition. Code and models are also released for research purposes\footnote{\url{https://github.com/stoa-xh91/ACFL}}.
\end{enumerate}  

\section{Related Work}
In this section, we review previous works that are strongly related to ours, including skeleton-based action recognition and knowledge distillation.
\subsection{Skeleton-based Action Recognition}
To date, recent skeleton-based action recognition methods have achieved great progress and can be roughly divided into three categories: 1) Recurrent Neural Network (RNN) based methods~\cite{Global_lstm,LSTM_skeleton}, 2) Convolutional Neural Network (CNN) based methods~\cite{PoseC3D} and 3) Graph Convolutional Network (GCN) based methods~\cite{ST-GCN_AAAI,Structural_gcn,MS_G3D,CTR_ICCV2021,shift_GCN}. 
Among them, the GCN-based approaches become the mainstream solutions as GCN-like architectures have a great potential to model structural data (e.g., skeleton sequences). Yan \etal~\cite{ST-GCN_AAAI} firstly propose the spatial-temporal Graph convolutional network (ST-GCN) for modeling spatial-temporal correlations between human joints, where a static topology based on human body structure is pre-defined to pass messages between joints in GCN. However, this static topology is fixed during inference, significantly limiting the expression power of GCNs. Then, many studies have been proposed to improve ST-GCN via deliberate designs of dynamic topologies. For example, Li \etal~\cite{Structural_gcn} design an encoder-decoder structure to capture joint correlations, while Liu \etal~\cite{MS_G3D} propose to dynamically generate topologies with a self-attention mechanism. Furthermore, Chen \etal~\cite{CTR_ICCV2021} explore non-shared topology, where the GCN model is forced to aggregate skeleton features in different channels with different dynamic topologies. In a different line of these works, Cheng \etal~\cite{shift_GCN} focus on an elaborate design of light-weight GCN by replacing heavy regular graph convolutions with spatial shift graph operations and point-wise convolutions, resulting in 10$\times$ less computational complexity.

Although significant progress has been achieved via elaborate designs of GCNs, they usually learn action correlated representation from single-form skeletons. To leverage multiple forms of skeletons with complementary cues, Shi \etal~\cite{2s-gcn} propose to construct two-stream GCNs, where action representations are generated by two different GCN models via separately taking as input joint and bone skeletons. Instead of adopting a two-stream paradigm which is computationally expensive, Tu \etal~\cite{joint2bone} adopt an early fusion strategy, which introduces negligible computational cost but hurts the unique representation of single-form skeletons. Furthermore, Song \etal~\cite{ResGCN} propose to utilize a residual GCN with a part-attention mechanism to generate complementary representations. Although impressive, they require the co-existence of all forms of skeletons during both training and inference stages, failing to handle the existence of only partial forms during inference. Different from them, our method focuses on the learning paradigm and empowers GCN models to generate complementary representations from single-form skeletons, which facilitates action recognition without model changes or extra input during the inference stage.

\subsection{Knowledge Distillation}
Numerous works~\cite{Learning_from_multiple_experts,Student_customized_KD,KD_Mimicking_Features, WangKTN++} for visual recognition have shown that knowledge distillation (KD) is an effective learning paradigm for exploiting the potential of deep neural networks without changing model architecture. This technique is often characterized by the so-called ``Student-Teacher'' learning framework, which is often computationally expensive as it requires a pre-trained ``teacher'' model with high-capacity. In KD, a student model with low-capacity is forced to approximate representations provided by the teacher model, thus reducing the gap between ``teacher'' and ``student''. However, general KD relies on a pre-trained large model, while a typical situation is that we do not always have a such powerful teacher model. To tackle this, a new variant of knowledge distillation based on collaborative learning is proposed in~\cite{Collaborative_learning}, which demonstrates that models with similar capacity can benefit each other through collaborative interaction. Our work is partially inspired by these works. Unlike those that enforce a model to fully mimic other representations or require a pre-trained large model, the proposed ACFL guides a model to learn useful representations from various models, which focuses on representation transfer among different models. Besides, how to transfer useful representations across various single-form skeletons based models, still remains an open question. As a supplement to previous works, our method can be viewed as an early attempt to explore representation transfer in the area of skeleton-based action recognition.

\section{Proposed Method}

\begin{figure*}[t]
	\centering
	\includegraphics[width=0.95\linewidth]{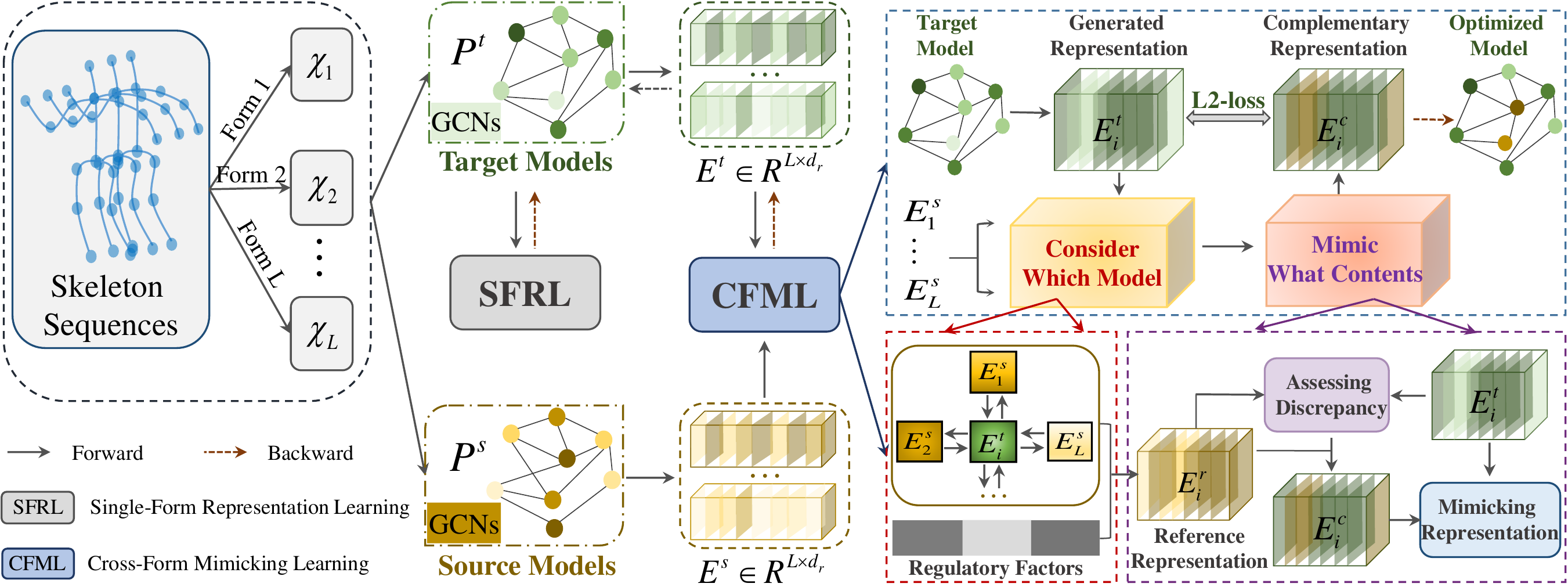}
	\caption{The overview of our proposed Adaptive Cross-Form Learning (ACFL). Given two groups of models, including target models that need to be optimized and source models that produce various representations from different single-form skeletons, the ACFL is applied to target models for facilitating action recognition. Specifically, the ACFL adopts Single-Form Representation Learning (SFRL) to force each target model to learn the single-form representation. And meanwhile, it also applies Cross-Form Mimicking Learning (CFML) to guide each target model to adaptively mimic useful representations from various single-form models.}
	\label{fig:framework}
\end{figure*}

\subsection{Preliminaries}
\noindent\textbf{Notations.} Formally, we denote different forms of input skeleton sequences as $\mathcal{X} \in \mathbb{R}^{M \times T \times V \times C}$, where $M$ is the number of persons, $T$ is the number of video frames, $V$ is the number of pre-detected skeleton points and $C$ indicates the dimensions of input skeleton. As an initial step, a learnable model is used to project the specific skeleton sequences into a semantic feature. In particular, the learnable model is denoted as $P:\mathbb{R}^{M \times T \times V \times C}\mapsto \mathbb{R}^{d}$, which is implemented by a GCN-like backbone network~\cite{ST-GCN_AAAI,shift_GCN,MS_G3D,CTR_ICCV2021}. Then, the semantic feature is formalized by Equ.~\ref{equ.ke}:
\begin{equation}
\begin{array}{lll}
{f} = P(\mathcal{X})
\end{array}
\label{equ.ke}
\end{equation}
where $f$ is a $(d)$-dimensional vector that encodes specific action semantic. Next, a classifier $\varPsi(\cdot)$ is used to project the semantic feature $f$ to a $N$-way categorical map $k \in \mathbb{R}^{N}$ for action recognition.  

\noindent\textbf{Learning Objective.} To enable model making correct recognition, existing methods often employ a standard cross-entropy loss as the optimization objective, which is formulated as Equ.~\ref{equ.standard_loss}:
\begin{equation}
\ell_{s}(k, y) = -\sum_{i=1}^{N}y_i \log(k_i) + (1-y_i)\log(1-k_i)
\label{equ.standard_loss}
\end{equation}
where $y$ is the ground-truth labels. It is worth noting that the semantic feature $f$ and categorical map $k$ are continually updated during training so that they become more discriminative from time to time. Therefore, the $f$ as well as $k$ can be used to express action correlated representations, which are derived from one specific form of skeleton sequences.

\subsection{Adaptive Cross-Form Learning}
\label{sec.method.main}
In this section, we introduce our proposed Adaptive Cross-Form Learning (ACFL) for skeleton-based action recognition. As illustrated in Fig.~\ref{fig:framework}, the ACFL divides models into two groups: 1) target models that need to be optimized; and 2) source models that produce various action representations derived from different forms of skeletons. Formally, we denote target models as $P^{t} = \{P^t_1,..., P^t_L\}$ and source models as $P^{s} = \{P^s_1,..., P^s_L\}$, where $L$ is the number of skeleton forms. Then, the ACFL applies two learning objectives to optimize each target model, including single-form representation learning (SFRL) and cross-form mimicking learning (CFML). The goal of the former one is to force target models to learn single-form representations, which is instantiated by Equ.~\ref{equ.standard_loss}. The latter one is utilized to force target models to adaptively mimic representations derived from various skeleton forms, which is the core of our ACFL. Next, we present details of the cross-form mimicking learning.

\noindent\textbf{Cross-Form Mimicking Learning.} For clarity, let us first denote the representations derived from source models as $E^s \in \mathbb{R}^{L\times d_r}$, where each representation is of size $d_r$. Similarly, we denote the representations generated by target models as $E^t \in \mathbb{R}^{L\times d_r}$. To produce complementary representations, we need to decide which source models should be considered and what useful contents will be mimicked. To achieve this, we generate reference representations $E^r \in \mathbb{R}^{L\times d_r}$ for target models by Equ.~\ref{equ.ke_ref}:
\begin{equation}
\begin{array}{lll}
A &= \sigma(\frac{(W_{q}{E^t}^{T})^{T}(W_{k}{E^s}^{T})}{\sqrt{d_r}}) \\
E^r &= (A \otimes \beta) E^s
\label{equ.ke_ref}
\end{array}
\end{equation}
where $\sigma(\cdot)$ is the softmax normalization. $W_{q}\in \mathbb{R}^{d_r\times d_r}$ and $W_{k}\in \mathbb{R}^{d_r\times d_r}$ are two learnable linear projection parameters. $A \in \mathbb{R}^{L\times L}$ is the importance weights that indicate which source model should be considered for each target model. $\otimes$ denotes broadcast multiplication. The $\beta \in \mathbb{R}^{1 \times L} $ is the regulatory factors, which re-adjust the importance weights in $A$. In particular, each element in $\beta$ is a pre-calculated accuracy score, which is a prior indicating the quality of representations derived from source models. In this way, deciding which source model as the reference not only depends on the relation between source models and target models, but also relies on the quality of representations that come from source models.

To decide what useful contents in reference representations for target models, we propose to measure the importance of contents by assessing the discrepancy between generated representations $E^t$ and reference representations $E^r$, as formulated in Equ.~\ref{equ.ke_diff}:
\begin{equation}
\begin{array}{lll}
Z &= \phi(W_v(E^t - E^r)^T) \\
E^c &= Z\odot E^r
\label{equ.ke_diff}
\end{array}
\end{equation}
where $\phi(\cdot)$ is the standard sigmoid non-linearly function that squashes input element into a range of (0, 1). $W_v\in \mathbb{R}^{d_r\times d_r}$ is the learnable linear projection parameters. $\odot$ is the element-wise multiplication. The $Z \in \mathbb{R}^{L\times d_r}$ is a weighting matrix and $E^c$ denotes the complementary representations that target models need to mimic. Note that each element in $Z$ decides whether corresponding content in reference representations should be mimicked or not. Next, we force each target GCN model to mimic complementary representations by directly reducing the difference between $E^c$ and $E^t$. Specifically, we use the L2 loss as formulated in Equ.~\ref{equ.distill_loss}:
\begin{equation}
\ell_{d}(E^c, E^t) = ||E^c - E^t||^2_2 
\label{equ.distill_loss}
\end{equation}
With standard learning objective and cross-form mimicking learning objective, the entire learning objective for target models can be re-written as Equ.~\ref{equ.model_loss}: 
\begin{equation}
\mathcal{L} = \frac{1}{L}\sum_{i=1}^{L}\ell_{s}(k^t_{i}, y) + \ell_{d}(E^c_i, E^t_i)
\label{equ.model_loss}
\end{equation}
Based on Equ.~\ref{equ.model_loss}, each target model not only learns representation from specific single-form input, but also learns to adaptively mimic useful representations produced by source models. 

\noindent\textbf{Instantiation.} In this work, we apply the proposed Adaptive Cross-Form Learning (ACFL) to train several GCN models for skeleton-based action recognition. Specifically, we consider three input forms (i.e., $L=3$), involving $joint$, $bone$ and both of $joint$ and $bone$. Following previous works~\cite{Student_customized_KD,Collaborative_learning}, semantic features $f$ and categorical activation maps $k$ are used for instantiation of action correlated representations (i.e., $E^{t}$ or $E^{s}$). Therefore, we consider two versions of ACFL: On-line ACFL and Off-line ACFL. In on-line version, target models and source models are identical, where representations $E^s$ used for cross-form mimicking learning are produced by target models. In this way, the source models are totally removed but the representations $E^s$ are poor in the early training phase. As for the off-line version, the source models are pre-trained with single-form representation learning, and then they are frozen when applying ACFL. Therefore, the quality of the $E^s$ is stably high in off-line version but requires extra pre-trained models. The effect of different kinds of data forms and the effect of two instantiation versions will be discussed in Section \ref{sec.exp.abla}.

\section{Experiments}

\begin{table*}[ht!]
	\centering
	\caption{Comparisons with representative GCNs on three challenging benchmarks, involving NTU-RGB+D 120, NTU-RGB+D 60 and UAV-Human. $M$ denotes million. The symbol ``$\ast$'' means that model is a re-implemented version. The ACFL denotes the proposed Adaptive Cross-Form Learning.}
	\setlength{\belowcaptionskip}{-0.5cm}%
	\resizebox{0.95\linewidth}{!}{
		\begin{tabular}{c|cc|cc|cc|c}
			\hline
			\multirow{2}{*}{Method}  & \multirow{2}{*}{Input} & \multirow{2}{*}{\#\textit{Param}} & \multicolumn{2}{c|}{NTU-RGB+D 120}  & \multicolumn{2}{c|}{NTU-RGB+D 60}   & UAV-Human         \\ \cline{4-8} 
			&                                  &         & \multicolumn{1}{c|}{X-Sub (\%)} & X-Set (\%)       & \multicolumn{1}{c|}{X-Sub (\%)}  & X-View (\%)  & \multicolumn{1}{c}{X-Sub (\%)} \\ \hline
			Shift-GCN~\cite{shift_GCN}      &   joint          & 0.71M    & \multicolumn{1}{c}{80.9}       & 83.2 & \multicolumn{1}{c}{87.8}       &  95.1             & -             \\
			Shift-GCN~\cite{shift_GCN}   &    bone            & 0.71M     & -       & -       &     -        & -       &   -  \\
			MS-G3D~\cite{MS_G3D}     &   joint          & 3.22M     & \multicolumn{1}{c}{-}       & -  & \multicolumn{1}{c}{89.4}       &  95.0             & -              \\
			MS-G3D~\cite{MS_G3D}   &    bone            & 3.22M     & \multicolumn{1}{c}{-}       & -   & \multicolumn{1}{c}{90.1}       &  95.3         & -            \\
			CTR-GCN~\cite{CTR_ICCV2021}      &   joint          & 1.46M   & \multicolumn{1}{c}{84.9}       & -   & \multicolumn{1}{c}{-}       & -                & -           \\
			CTR-GCN~\cite{CTR_ICCV2021}   &    bone            & 1.46M     & \multicolumn{1}{c}{85.7}       & 87.4   & \multicolumn{1}{c}{-}       &     -        & -             \\	
			\hline
			\hline
			Shift-GCN$^{\ast}$     &   joint          & 0.71M    & \multicolumn{1}{c}{82.8}       & 84.3  & \multicolumn{1}{c}{88.0}       &  93.6       &    41.3          \\
			Shift-GCN$^{\ast}$ w \textbf{ACFL}     &   joint          & 0.71M    & \multicolumn{1}{c}{\textbf{85.1$^{\mathbf{+2.3}}$}}       & \textbf{85.9$^{\mathbf{+1.6}}$}    & \multicolumn{1}{c}{\textbf{90.0$^{\mathbf{+2.0}}$}}       &  \textbf{94.1$^{\mathbf{+0.5}}$}     &   \textbf{43.2$^{\mathbf{+1.9}}$}           \\
			Shift-GCN$^{\ast}$     &    bone            & 0.71M   & \multicolumn{1}{c}{83.7}     & 84.2   & \multicolumn{1}{c}{89.6}       &   92.4        &  40.5       \\
			Shift-GCN$^{\ast}$ w \textbf{ACFL}           &    bone            & 0.71M   & \multicolumn{1}{c}{\textbf{85.5$^{\mathbf{+1.8}}$}}     & \textbf{86.7$^{\mathbf{+2.5}}$}     & \multicolumn{1}{c}{\textbf{90.3$^{\mathbf{+0.7}}$}}  &  \textbf{93.9$^{\mathbf{+1.5}}$}       & \textbf{41.6$^{\mathbf{+1.1}}$}        \\
			Shift-GCN$^{\ast}$    &    joint \& bone       & 0.71M  & \multicolumn{1}{c}{85.4}     & 86.2   & \multicolumn{1}{c}{89.5}       &   93.4          &  41.2         \\
			Shift-GCN$^{\ast}$ w \textbf{ACFL}    &    joint \& bone      & 0.71M   & \multicolumn{1}{c}{\textbf{86.9$^{\mathbf{+1.5}}$}}     & \textbf{87.8$^{\mathbf{+1.6}}$}  & \multicolumn{1}{c}{\textbf{90.5$^{\mathbf{+1.0}}$}}       &   \textbf{94.5$^{\mathbf{+1.1}}$}          & \textbf{42.6$^{\mathbf{+1.4}}$}          \\
			\hline
			\hline
			MS-G3D$^{\ast}$     &   joint    & 3.22M    & \multicolumn{1}{c}{85.4}       & 87.2 & \multicolumn{1}{c}{89.7}       &  94.7            &   42.2          \\
			MS-G3D$^{\ast}$ w \textbf{ACFL}  &   joint   & 3.22M      & \multicolumn{1}{c}{\textbf{87.3$^{\mathbf{+1.9}}$}}       & \textbf{88.7$^{\mathbf{+1.5}}$}   & \multicolumn{1}{c}{\textbf{90.3$^{\mathbf{+0.6}}$} }       &     \textbf{94.7$^{\mathbf{+0.0}}$}        &    \textbf{43.3$^{\mathbf{+1.1}}$}       \\
			MS-G3D$^{\ast}$     &    bone            & 3.22M   & \multicolumn{1}{c}{86.7}       & 87.6     & \multicolumn{1}{c}{90.5}       &  94.5             &   41.7        \\
			MS-G3D$^{\ast}$ w \textbf{ACFL}   &  bone  & 3.22M    & \multicolumn{1}{c}{\textbf{88.1$^{\mathbf{+1.4}}$}}       &    \textbf{89.0$^{\mathbf{+1.4}}$}   & \multicolumn{1}{c}{\textbf{91.0$^{\mathbf{+0.5}}$}}       &  \textbf{95.5$^{\mathbf{+1.0}}$}         &   \textbf{43.3$^{\mathbf{+1.6}}$}          \\
			\hline
			\hline
			CTR-GCN$^{\ast}$   &   joint      & 1.46M    & \multicolumn{1}{c}{84.9}       & 86.5 & \multicolumn{1}{c}{89.6}       & 94.5  & 41.7              \\
			CTR-GCN$^{\ast}$ w \textbf{ACFL}   &   joint   & 1.46M   & \multicolumn{1}{c}{\textbf{87.3$^{\mathbf{+2.4}}$}}    & \textbf{88.7$^{\mathbf{+2.2}}$} & \multicolumn{1}{c}{\textbf{91.2$^{\mathbf{+1.6}}$}}    & \textbf{96.4$^{\mathbf{+1.9}}$}   &  \textbf{43.8$^{\mathbf{+2.1}}$}           \\
			CTR-GCN$^{\ast}$  &    bone    & 1.46M   & \multicolumn{1}{c}{85.7}       & 87.4 & \multicolumn{1}{c}{90.2}        &  94.9        & 41.0            \\
			CTR-GCN$^{\ast}$ w \textbf{ACFL}   &    bone    & 1.46M & \multicolumn{1}{c}{\textbf{88.4$^{\mathbf{+2.7}}$}}   & \textbf{89.5$^{\mathbf{+2.1}}$}  & \multicolumn{1}{c}{\textbf{91.4$^{\mathbf{+1.2}}$}} &  \textbf{96.4$^{\mathbf{+1.5}}$}    &     \textbf{43.3$^{\mathbf{+2.3}}$}         \\
			CTR-GCN$^{\ast}$ &    joint \& bone     & 1.46M   & \multicolumn{1}{c}{86.9}   & 88.8   & \multicolumn{1}{c}{90.9}    &  95.6    & 42.4     \\			
			CTR-GCN$^{\ast}$ w \textbf{ACFL}   &    joint \& bone     & 1.46M  & \multicolumn{1}{c}{\textbf{89.3$^{\mathbf{+2.4}}$}}  & \textbf{90.2$^{\mathbf{+1.4}}$} & \multicolumn{1}{c}{\textbf{92.0$^{\mathbf{+1.1}}$}}    &    \textbf{96.6$^{\mathbf{+1.0}}$}      &  \textbf{44.2$^{\mathbf{+1.8}}$}            \\
			\hline
		\end{tabular}
	}
	\label{tab:main_res}
\end{table*}
\subsection{Datasets}
We evaluate our method on following three challenging benchmarks~\cite{Ntu60,NTU120,UAV-human}:

\noindent\textbf{NTU-RGB+D 60}~\cite{Ntu60}. This dataset is the most widely used dataset for skeleton-based action recognition. It contains 56,880 skeleton sequence samples that involve 60 action categories performed by 40 subjects. In addition, each human skeleton is represented by 25 joints with 3D coordinates captured by 3 cameras from different horizontal angles. Following official settings, two splits are used: 1) Cross-Subject (X-sub): 20 subjects are used for training, and the rest for testing. 2) Cross-View (X-view): skeleton sequences collected from camera 1 are used for testing and the rest are used for training. 

\noindent\textbf{NTU-RGB+D 120}~\cite{NTU120}. This dataset is an extended version of NTU-RGB+D 60 dataset. It contains 114,480 skeleton sequence samples and involves 120 action classes performed by 106 subjects in 155 viewpoints and 3 different camera perspectives. Two splits are used: 1) Cross-Subject (X-Sub): 106 subjects are separately split into the training set and testing set, where each set contains 53 subjects; 2) Cross-Setup (X-Set): samples with even setup IDs are used for training and the rest for testing.  

\noindent\textbf{UAV-Human}~\cite{UAV-human}. This dataset is recently released for human behavior understanding with unmanned aerial vehicles \cite{9583266}, including 22,476x3 multi-modal video sequences. It involves 155 different activity categories performed by 119 distinct subjects in 45 different environment sites. The dataset is collected by a flying UAV in multiple urban and rural districts in both daytime and nighttime, hence covering extensive diversities w.r.t subjects, backgrounds, illuminations, weathers, occlusions, camera motions, and UAVs flying attitudes. In this dataset, 17 body joints are used to represent each person. One split (X-Sub) is used: 89 subjects for training and the remaining 30 subjects for testing.

\noindent\textbf{Evaluation metric.} We follow the official evaluation protocols of each dataset for fair comparisons and report standard mean average accuracy on the corresponding test set.

\subsection{Implementation details} 
To be fair, all models are implemented with Pytorch and all experiments are conducted on a server with 8 NVIDIA RTX GPUs (24GB memory per-card). We train GCNs (16 skeleton sequences per GPU) for 65 epochs with an initial learning rate of 0.1, and respectively drop it with a factor 0.1 at the 35-th and 55-th epoch.
For data pre-processing, each sequencing sample is resized to 64 frames. Other details are identical to CTR-GCN~\cite{CTR_ICCV2021}.

\subsection{Main Results}
In this work, three representative GCN models, i.e., Shift-GCN~\cite{shift_GCN}, MS-G3D~\cite{MS_G3D} and CTR-GCN~\cite{CTR_ICCV2021}, are used as baseline models for investigating the effectiveness of our proposed ACFL. In particular, Shift-GCN is a light-weight model while MS-G3D is with the highest model capacity. All baseline models are optimized via single-form representation learning based on a cross-entropy loss only (i.e., $\ell_{s}$). For fair comparisons with corresponding baselines, we report our re-implemented results of these models, which are generally higher than or comparable with that are reported in papers. 

The experimental results are summarized in Tab.~\ref{tab:main_res}. With the help of our proposed ACFL, the CTR-GCN that takes as input both joint and bone achieves the best performance, where it respectively obtains 89.3\% accuracy on NTU-RGB+120 (X-Sub), 90.2\% on NTU-RGB+120 (X-Set), 92.0\% on NTU-RGB+60 (X-Sub), 96.6\% on NTU-RGB+60 (X-View) and 44.2\% on UAV-Human (X-Sub). In addition, the experimental results can be further summarized into the following conclusions: (1) The potential of single-form GCNs is further exploited by ACFL, since all baselines learned from different skeleton forms are improved by stable performance gains ranging from 1.0\% to 2.0\%. This indicates the generalizability and effectiveness of the proposed ACFL. (2) Single-form GCN models (either joint based or bone based) trained with ACFL are generally better than or comparable with corresponding hybrid-form based baseline models (i.e., simultaneously feeding joint and bone). For example, either joint or bone based CTR-GCN trained with ACFL outperforms hybrid-form baseline CTR-GCN, i.e., 87.3\% vs 86.9\% and 88.4\% vs 86.9\%. This suggests mimicking various single-form models results in better action representations, comparing with directly feeding multiple forms of skeletons into one model. (3) The effectiveness of ACFL depends on the scale of training data, where the performance gains on large datasets (i.e., around 2.0\% on NTU-RGB+D 120) are often higher than that of small datasets (i.e., around 1.1\% on NTU-RGB+D 60 and around 1.6\% on UAV-Human). (4) The ACFL is robust against the model capacity, since it brings relatively stable performance gains regardless of the model capacity.

\begin{table}[t]
	\centering
	\caption{Ablation studies on learning paradigm. J-CTR-GCN means CTR-GCN model takes as input joint-form skeletons while B-CTR-GCN means CTR-GCN model takes as input bone-form skeletons. The BJ-CTR-GCN means that CTR-GCN model is based on hybrid-form of skeletons (i.e., joint and bone together).}
	\setlength{\belowcaptionskip}{-0.5cm}%
	\renewcommand\arraystretch{1.2}
	\resizebox{0.90\linewidth}{!}{
		\begin{tabular}{c|cc|c}
			\hline
			\multirow{2}{*}{Target Model}       &  \multicolumn{2}{c|}{Adaptive Cross-Form Learning}   &   \multirow{2}{*}{Acc (\%)}      \\ \cline{2-3}
			& \multicolumn{1}{l}{On-line ACFL}   &  \multicolumn{1}{c|}{Off-line ACFL}  & \\ 
			\hline
			\multirow{3}{*}{J-CTR-GCN}  & - & - & 84.9    \\ 
			& \checkmark &  &  86.4$^{\mathbf{+1.5}}$     \\ 
			&  &  \checkmark &  87.3$^{\mathbf{+2.4}}$    \\ \hline
			\multirow{3}{*}{B-CTR-GCN}   & - & - & 85.7    \\ 
			& \checkmark &   &  87.6$^{\mathbf{+1.9}}$    \\ 
			&  & \checkmark & 88.4$^{\mathbf{+2.7}}$     \\ \hline
			\multirow{3}{*}{BJ-CTR-GCN}  & - & - & 86.9    \\ 
			& \checkmark &   & 88.7$^{\mathbf{+1.8}}$     \\ 
			&  & \checkmark & 89.3$^{\mathbf{+2.4}}$     \\ 
			\hline
		\end{tabular}
	}
	\label{tab:version_comp}
\end{table}

\begin{table}[t]
	\centering
	\caption{Ablation studies on cross-form mimicking learning. Investigating the effect of three source models, where each source model produces action representation derived from single-form skeletons, e.g., joint, bone or both of joint and bone.}
	\vspace{-0.em}
	\setlength{\belowcaptionskip}{-0.5cm}%
	\renewcommand\arraystretch{1.2}
	\resizebox{1.0\linewidth}{!}{
		\begin{tabular}{c|ccc|c}
			
			\hline
			\multirow{2}{*}{Target model}       &  \multicolumn{3}{c|}{Source Model}   &   \multirow{2}{*}{Acc (\%)}      \\ \cline{2-4}
			& \multicolumn{1}{c|}{J-CTR-GCN}   & \multicolumn{1}{c|}{B-CTR-GCN} & \multicolumn{1}{c|}{BJ-CTR-GCN} & \\ 
			\hline
			\hline
			\multirow{6}{*}{J-CTR-GCN}   & - & - & -  &  84.9 \\\cline{2-5}
			& \checkmark &  & &  86.1$^{\mathbf{+1.2}}$   \\
			&  & \checkmark &   &  87.4$^{\mathbf{+2.5}}$   \\
			&  &  &  \checkmark &  87.0$^{\mathbf{+2.1}}$ \\
			& \checkmark & \checkmark  &   & 87.1$^{\mathbf{+2.2}}$  \\
			& \checkmark &  \checkmark & \checkmark  & 87.3$^{\mathbf{+2.4}}$  \\
			\hline
			\hline
			\multirow{6}{*}{B-CTR-GCN}   & - & - &-   & 85.7  \\\cline{2-5}
			& \checkmark &  & &  86.5$^{\mathbf{+0.8}}$    \\
			&  & \checkmark &   &  87.8$^{\mathbf{+2.1}}$   \\
			&  &  & \checkmark  &  87.6$^{\mathbf{+1.9}}$ \\
			& \checkmark & \checkmark &   &  87.5$^{\mathbf{+1.8}}$   \\
			& \checkmark &  \checkmark & \checkmark  & 88.4$^{\mathbf{+2.7}}$  \\
			\hline
		\end{tabular}
	}
	\label{tab:abla_study_peer_kind}
\end{table}

\begin{table}[t]
	\centering
	\caption{Ablation studies on Action Representation. Investigating the effect of two types of action representation in ACFL, including semantic feature and categorical map.}
	\vspace{-0.em}
	\setlength{\belowcaptionskip}{-0.5cm}%
	\renewcommand\arraystretch{1.2}
	\resizebox{0.99\linewidth}{!}{
		\begin{tabular}{c|cc|c}
			\hline
			\multirow{2}{*}{Target Model}       &  \multicolumn{2}{c|}{Action Representations}   &   \multirow{2}{*}{Acc (\%)}      \\ \cline{2-3}
			& \multicolumn{1}{l}{Semantic Feature}   &  \multicolumn{1}{c|}{Categorical Map}  & \\ 
			\hline
			\hline
			\multirow{4}{*}{J-CTR-GCN}   & \quad- & - & 84.9    \\ \cline{2-4}
			& \quad\checkmark & - &  85.4$^{\mathbf{+0.5}}$   \\
			& \quad- & \checkmark &  86.8$^{\mathbf{+1.9}}$  \\
			& \quad\checkmark &  \checkmark  & 87.3$^{\mathbf{+2.4}}$  \\
			\hline
		\end{tabular}
	}
	\label{tab:abla_study_feat_logit}
\end{table}

\begin{table}[t]
	\centering
	\caption{Ablation studies under cross-architecture setting. Investigating the generalizability of the proposed method by applying ACFL among heterogeneous or homogeneous GCNs.}
	\vspace{-0.em}
	\setlength{\belowcaptionskip}{-0.5cm}%
	\renewcommand\arraystretch{1.2}
	\resizebox{0.95\linewidth}{!}{
		\begin{tabular}{c|c|c|c|c}
			\hline
			\multirow{2}{*}{Target Model}       &  \multicolumn{3}{c|}{Source Model}   &   \multirow{2}{*}{Acc (\%)}      \\ \cline{2-4}
			& \multicolumn{1}{c|}{\textit{Shift-GCN}}   & \multicolumn{1}{c|}{\textit{CTR-GCN}} & \multicolumn{1}{c|}{\textit{MS-G3D}} & \\ 
			\hline
			\hline
			\multirow{4}{*}{J-CTR-GCN}   & - & - & -  &  84.9 \\\cline{2-5}
			& \checkmark &  &   & 86.4$^{\mathbf{+1.5}}$   \\
			&  & \checkmark &   &  87.3$^{\mathbf{+2.4}}$  \\
			&  &   & \checkmark  & 87.0$^{\mathbf{+2.1}}$    \\
			\hline
			\hline
			\multirow{4}{*}{B-CTR-GCN}   & - & - & -  &  85.7 \\\cline{2-5}
			& \checkmark &  &   & 87.5$^{\mathbf{+1.8}}$    \\
			&  & \checkmark &   &  88.4$^{\mathbf{+2.7}}$ \\
			&  &   & \checkmark  & 87.5$^{\mathbf{+1.8}}$    \\
			\hline
		\end{tabular}
	}
	\label{tab:abla_study_cross_model}
\end{table}

\begin{table*}[t]
	\centering
	\caption{Quantitative Analysis. Investigating the effect of proposed method for recognition of top-10 ``hard'' action classes.}
	\setlength{\belowcaptionskip}{-0.5cm}%
	\renewcommand\arraystretch{1.2}
	\resizebox{1.0\linewidth}{!}{
		\begin{tabular}{c|c|c|c|c|c|c|c|c|c|c|c}
			\hline                                                                       
			Setting & drink water                  & reading           & make phone                 & touch neck          & juggling balls          & grab stuff & thumb down & ok sign & stable book & cutting nails & Mean       \\ \hline
			J-CTR-GCN                & 69\%                   & 53\%          & \textbf{61\%}                   & \textbf{84\%}          & 78\%          & 70\%   &  53\%  &  35\% & 57\%  & 61\%  & 62\%\\
			B-CTR-GCN                & \textbf{76\%}                    & 53\%          & 58\%                  & 79\%          & 80\%          & \textbf{81\%} & \textbf{70\%}    &  61\% & 55\%  &  \textbf{68\%}  & 68\%\\
			BJ-CTR-GCN               & 73\%                   & \textbf{61\%}          & 60\%                 & 83\%          & \textbf{83\%}          & 79\%  &  68\%  & \textbf{66\%} & \textbf{58\%}  & 62\%  & \textbf{69\%} \\ \hline
			J-CTR-GCN w ACFL         & 76\%$^{\mathbf{+7}}$                   & 63\%$^{\mathbf{+10}}$          & 63\%$^{\mathbf{+2}}$           & \textbf{87\%}$^{\mathbf{+3}}$          & 82\%$^{\mathbf{+4}}$          & 77\%$^{\mathbf{+7}}$     & 50\%$^{\mathbf{-3}}$  & 45\%$^{\mathbf{+10}}$ & 61\%$^{\mathbf{+4}}$  & 67\%$^{\mathbf{+6}}$   & 67\%$^{\mathbf{+5}}$ \\
			B-CTR-GCN w ACFL         & \textbf{77\%}$^{\mathbf{+1}}$                    & 60\%$^{\mathbf{+7}}$          & 64\%$^{\mathbf{+6}}$              & 86\%$^{\mathbf{+7}}$          & \textbf{85\%}$^{\mathbf{+5}}$          & 82\%$^{\mathbf{+1}}$     & \textbf{73\%}$^{\mathbf{+3}}$  & 61\%$^{\mathbf{+0}}$ &  \textbf{64\%}$^{\mathbf{+9}}$   & 66\%$^{\mathbf{-2}}$   & 72\%$^{\mathbf{+4}}$ \\
			BJ-CTR-GCN w ACFL        & 75\%$^{\mathbf{+2}}$                   & \textbf{65\%}$^{\mathbf{+4}}$          & \textbf{65\%}$^{\mathbf{+5}}$           & \textbf{87\%}$^{\mathbf{+4}}$          & 84\%$^{\mathbf{+1}}$          & \textbf{83\%}$^{\mathbf{+4}}$     & 69\%$^{\mathbf{+1}}$  & \textbf{70\%}$^{\mathbf{+4}}$ & 63\%$^{\mathbf{+5}}$   &  \textbf{68\%}$^{\mathbf{+6}}$   &  \textbf{73\%}$^{\mathbf{+4}}$ \\ \hline 
		\end{tabular}
	}
	\label{tab:abla_study_hard}
\end{table*}

\begin{table}[t]
	\centering
	\caption{Classification accuracy comparison with state-of-the-art methods on the NTU RGB+D 120 dataset.}
	\setlength{\belowcaptionskip}{-0.5cm}%
	\renewcommand\arraystretch{1.2}
	\resizebox{0.80\linewidth}{!}{
		\begin{tabular}{c|c|c}
			\hline
			\multicolumn{1}{c|}{\multirow{2}{*}{Methods}} & \multicolumn{2}{c}{NTU-RGB+D 120} \\ \cline{2-3}
			& X-Sub (\%)      & X-Set (\%)      \\ 
			
			
			\hline
			2s-AGCN~\cite{2s-gcn}                          &   82.9    &  84.9     \\ 
			SGN~\cite{SGN}                                 &   79.2    &  81.5     \\    
			DC-GCN+ADG~\cite{DC_GCN+ADG}          &   86.5    &  88.1    \\ 
			PA-ResGCN-B19~\cite{ResGCN}             &   87.3    &  88.3    \\ 
			Dynamic GCN~\cite{Dynamic_GCN}                 &   87.3    &  88.6    \\ 
			\hline
			\hline
			Shift-GCN (4s)~\cite{shift_GCN}             &   85.9    &  87.6     \\ \hline
			ACFL-Shift-GCN (1s)   & 85.5   &  86.7      \\ 
			ACFL-Shift-GCN (2s)   & 86.9   &  88.0      \\  
			ACFL-Shift-GCN (3s)   & \textbf{87.5}   &  \textbf{88.5}    \\ 
			\hline
			\hline
			MS-G3D (4s)~\cite{MS_G3D}             &   86.9    &  88.4    \\\hline
			ACFL-MS-G3D (1s)   & 88.1 &  89.0       \\ 
			ACFL-MS-G3D (2s)   & \textbf{88.6} &  \textbf{89.8}       \\  
			\hline 
			\hline
			CTR-GCN (4s)~\cite{CTR_ICCV2021}                 &   88.9    &  90.6    \\\hline
			ACFL-CTR-GCN (1s)   & 88.4     &  89.5  \\  
			ACFL-CTR-GCN (2s)   & 89.1     &  90.5  \\  
			ACFL-CTR-GCN (3s)   & \textbf{89.7}     &  \textbf{90.9}  \\ 
			
			\hline
		\end{tabular}	
	}
	
	\label{tab:comp_sota_ntu120}
\end{table}

\begin{table}[t]
	\centering
	\caption{Classification accuracy comparison with state-of-the-art methods on the NTU RGB+D 60 dataset.}
	\vspace{-0.1em}
	\setlength{\belowcaptionskip}{-0.5cm}%
	\resizebox{0.80\linewidth}{!}{
		\begin{tabular}{c|c|c}
			\hline
			\multicolumn{1}{c|}{\multirow{2}{*}{Methods}} & \multicolumn{2}{c}{NTU-RGB+D 60}  \\ \cline{2-3}
			& X-Sub (\%)      & X-View (\%)      \\ 
			\hline
			ST-GCN~\cite{ST-GCN_AAAI}               &  81.5       &  88.3      \\   
			2s-AGCN~\cite{2s-gcn}                   &  88.5       &  95.1       \\
			AGC-LSTM~\cite{agc_lstm}                &  89.2       &  95.0      \\ 
			DGNN~\cite{DGNN}                        &  89.9       &  96.1       \\
			DC-GCN+ADG~\cite{DC_GCN+ADG}            &  90.8       &  96.6       \\   
			SGN~\cite{SGN}                          &  89.0       &  94.5        \\ 
			PA-ResGCN-B19~\cite{ResGCN}      &  90.9       &  96.0    \\ 
			Dynamic GCN~\cite{Dynamic_GCN}          &  91.5       &  96.0      \\ 
			\hline
			\hline
			Shift-GCN (4s)~\cite{shift_GCN}      &  90.7       &  96.5       \\ \hline 
			ACFL-Shift-GCN (1s)   & 90.3   &  93.9      \\ 
			ACFL-Shift-GCN (2s)   & 91.2   &  94.9      \\  
			ACFL-Shift-GCN (3s)   & \textbf{91.4}   &  \textbf{95.1}    \\ 
			\hline
			\hline
			MS-G3D (2s)~\cite{MS_G3D}      &  91.5       &  96.2       \\ \hline 
			ACFL-MS-G3D (1s)   & 91.0 &  95.5       \\ 
			ACFL-MS-G3D (2s)   & \textbf{91.4} &  \textbf{95.6}       \\  
			\hline
			\hline
			CTR-GCN (4s)~\cite{CTR_ICCV2021}          &  92.4       &  96.8       \\ \hline 
			ACFL-CTR-GCN (1s)   & 91.4     &  96.4  \\ 
			ACFL-CTR-GCN (2s)   & 92.1     &  97.0  \\  
			ACFL-CTR-GCN (3s)   & \textbf{92.5}     &  \textbf{97.1}  \\ 
			\hline
		\end{tabular}
	}
	\label{tab:comp_sota_ntu60}
\end{table}

\begin{table}[t]
	\centering
	\caption{Classification accuracy comparison with state-of-the-art methods on the UAV-Human dataset.}
	\setlength{\belowcaptionskip}{-0.5cm}%
	\resizebox{0.55\linewidth}{!}{
		\begin{tabular}{c|c}
			\hline
			\multicolumn{1}{c|}{Methods}           &  Acc (\%)           \\ \hline
			ST-GCN~\cite{ST-GCN_AAAI}           &  30.3       \\ 
			DGNN~\cite{DGNN}                    &  29.9       \\
			2s-AGCN~\cite{2s-gcn}               &  34.8       \\   
			HARD-Net~\cite{Hard-net}            &  37.0       \\
			\hline
			\hline
			Shift-GCN (2s)~\cite{shift_GCN} &  42.9       \\ \hline
			ACFL-Shift-GCN (2s)   & 43.0         \\  
			ACFL-Shift-GCN (3s)   & \textbf{43.4}         \\ 
			\hline
			\hline
			MS-G3D (2s)~\cite{MS_G3D} &  43.4      \\ \hline
			ACFL-MS-G3D (1s)   &  43.3       \\ 
			ACFL-MS-G3D (2s)   & \textbf{43.9}         \\  
			\hline
			\hline 
			CTR-GCN (2s)~\cite{CTR_ICCV2021}     &  43.4      \\ \hline
			ACFL-CTR-GCN (2s)   &   44.3      \\  
			ACFL-CTR-GCN (3s)   & \textbf{45.3}         \\ 
			\hline
		\end{tabular}
	}
	
	\label{tab:comp_sota_uav}
	\vspace{-1.0em}
\end{table}

\subsection{Ablation Studies}
\label{sec.exp.abla}
In this section, we focus on the investigation of the proposed method. In specific, we first investigate two options of the proposed ACFL. Next, we study the effect of different forms of skeletons. Then, we investigate the effect of different types of action representations. Further, we study the generalizability of our proposed method under two challenging settings. Finally, we conduct a detailed analysis of learned models. For fairness, all ablation experiments are conducted on NTU-RGB+D 120 (X-Sub) and the CTR-GCN is adopted as the baseline model, unless otherwise stated.

\noindent\textbf{The effect of instantiation of ACFL.} As described in Section~\ref{sec.method.main}, the proposed ACFL can be instantiated as two versions, i.e., On-line ACFL and Off-line ACFL. In this section, we would like to know which one is more beneficial for target models. Thus, we compare three different settings: 1) all target models are trained via single-form representation learning only; 2) all target models are trained via the On-line ACFL; and 3) all target models are trained by applying the Off-line ACFL. In particular, we adopt three CTR-GCNs as target models, where each one takes as input the single form of skeletons. For simplicity, we separately denote three models as J-CTR-GCN, B-CTR-GCN, and BJ-CTR-GCN, which are respectively based on joint-form skeletons, bone-form skeletons, and hybrid-form (i.e., joint and bone together) skeletons.

The experimental results are summarized in Tab.~\ref{tab:version_comp}. From the results, we observe that each target model is improved when applying proposed On-line ACFL or Off-line ACFL, obtaining around 2\% performance gains in terms of accuracy. This indicates the effectiveness of the proposed On-line ACFL as well as Off-line ACFL. Furthermore, the Off-line ACFL is more beneficial for target models than On-line ACFL, as it achieves slightly higher accuracy scores, i.e., 87.3\% vs 86.4\% for joint-based models, 88.4\% vs 87.6\% for bone-based models, 89.3\% vs 88.7\% for hybrid-form based models. We conjecture the possible reason is that the representations derived from source models are not discriminative enough as they need to be continually updated in On-line ACFL. Instead, representations derived from source models in Off-line ACFL are stable yet reliable in thorough training, thus benefiting target models in cross-form mimicking learning. Based on this observation, we adopt the Off-line ACFL as our method in the following experiments.

\noindent\textbf{Ablation studies of Cross-Form Mimicking Learning.} The core component of ACFL lies in cross-form mimicking learning, where each target model is forced to mimic complementary representations derived from various skeleton forms. Thus, one question is posed: how much each source form contributes to the final complementary representation. To answer this, we adopt two target models, which are separately based on $joint$ and $bone$ skeletons.

The experimental results are summarized in Tab.~\ref{tab:abla_study_peer_kind}. From the results, we first observe that each target model is improved even if we adopt one source model in ACFL. In particular, when the target model and source model are based on the same form of skeletons, applying the proposed ACFL also benefits the target model, obtaining stable performance gains (1.2\% for joint-based models and 2.1\% for bone-based models). This implies that two trained GCN models are supplement to each other in some cases, even if they are based on the same form of skeletons. Thus, one can conclude that the ACFL effectively forces the target model to strengthen what it has learned by mimicking the source model based on the same form of skeletons. Secondly, mimicking representations derived from two forms of skeletons improves the joint-based target model by 2.4\% accuracy, but the performance gains start to saturate when mimicking all source models. Similar gains for the bone-based target model are also observed. Consistent results indicate that representations learned in two different source models are approximately complimentary.  

\noindent\textbf{The effect of action representations for mimicking.} As reported in Tab.~\ref{tab:abla_study_feat_logit}, mining useful representations from semantic features or categorical maps contributes performance improvement. In particular, mimicking semantic features improves the CTR-GCN from 84.9\% to 85.4\% while mimicking categorical maps brings 1.9\% gains. Mimicking all two representations from both $f$ and $k$ achieves the best, which indicates the effectiveness of the proposed ACFL for various representation mining. 

\noindent\textbf{The generalizability of the ACFL.} Since the ACFL is model-agnostic, we investigate the generalization capability of the ACFL by applying it among heterogeneous or homogeneous GCNs. Specifically, the CTR-GCN is used as the target model while Shift-GCN, as well as MS-G3D, are used as source models. Under this setting, the CTR-GCN is forced to mimic representations produced by heterogeneous architectures. The experimental results are reported in Tab.~\ref{tab:abla_study_cross_model}. When adopting joint-based CTR-GCN as the baseline, we observe that the target model is improved after mimicking representations stored in heterogeneous architectures by ACFL, where the performance is boosted from 84.9\% to 86.4\% and the counterpart reaches 2.4\% when respectively setting Shift-GCN and MS-G3D as the source models. Similar improvements are observed when adopting bone-based CTR-GCN as the baseline. The consistent improvements demonstrate that the ACFL is effective in mining useful representations regardless of model architectures, showing good generalization of our proposed method.

\noindent\textbf{An analysis of learned model.} To further attain insight into the learned model, in Tab.~\ref{tab:abla_study_hard}, we report class-level accuracy scores produced by various CTR-GCN models. In particular, 10 classes that CTR-GCN performs worst are selected for evaluation. From the results, we have the following observations: 1) Different action prefers different forms of skeletons, which is in line with the motivation as stated in Section~\ref{sec:intro}. For example, head-related actions, such as \textit{make phone} and \textit{touch neck}, are well recognized by J-CTR-GCN. However, B-CTR-GCN shows better performance for recognizing hand-related actions like \textit{drink water}. 2) Directly feeding multiple forms of skeletons into one model is not always a good choice for generating complementary representations, since for some cases BJ-CTR-GCN shows lower performance (See second row, column 2, 4, 5, 7, 8 and 11) than that of J-CTR-GCN or B-CTR-GCN. We conjecture the possible reason behind this is that simply fusing two forms will hurt the unique representation of each form in GCN model. 3) The performance gap between all pairs of CTR-GCN models is significantly reduced after applying ACFL. This indicates that representations produced by CTR-GCN models are similar, suggesting that each model successfully mimics complementary action representations after applying the proposed ACFL. 4) For recognizing hard actions that are with minor changes, such as ``making ok sign'' or ``thumb down'', the CTR-GCNs perform worst even if they are trained via ACFL. This is because the change of skeleton point is quite minor, which makes it difficult to distinguish ``making ok sign'' from other hand-related actions. To precisely identify such actions, other cues such as RGB appearance or optical flow can also be considered. For more analysis of ablation studies, we refer readers to supplementary materials.

\subsection{Comparisons with the State-of-the-Arts}
In this section, we compare the GCNs that are trained with ACFL with other state-of-the-art methods on the NTU-RGB+D 120, NTU-RGB+D 60 and UAV-Human. Specifically, we use the Shift-GCN with ACFL, MS-G3D with ACFL and CTR-GCN with ACFL for comparison. For clarity, we respectively denote them as ACFL-Shift-GCN, ACFL-MS-G3D and ACFL-CTR-GCN. Following previous works that fuse results of different modalities, we adopt the same settings for fairness and build three settings: 1) 1 stream (1s), where prediction results come from bone-based GCN; 2) 2 streams (2s), where prediction results are obtained by fusing predictions from two GCNs ($joint$ and $bone$); 3) 3 streams (3s), where we fuse prediction results of three forms, i.e., $joint$, $bone$ and both of them.

Comparison results on NTU-RGB+D 120, NTU-RGB+D 60 and UAV-Human are respectively reported in Tab.~\ref{tab:comp_sota_ntu120}, Tab.~\ref{tab:comp_sota_ntu60} and Tab.\ref{tab:comp_sota_uav}. On these three datasets, our GCNs consistently outperform all existing methods and achieve a new record on all benchmarks. In particular, the ACFL-CTR-GCN achieves the best on all three datasets, where it achieves accuracy scores of 89.7\% and 90.9\% on NTU-RGB+D 120, 92.5\% and 97.1\% on NTU-RGB+D 60 and 45.3\% on UAV-Human. 

\section{Conclusion}
In this paper, we propose a novel learning paradigm named adaptive cross-form learning (ACFL), aiming to empower well-designed GCNs to learn complementary representation from single-form skeletons without changing model capacity. By adaptively mimicking useful action representations from other forms of skeletons, each single-form GCN model can smartly strengthen what it has learned, and thus can exploit the model potential and facilitate action recognition as well. Moreover, the proposed ACFL can be easily applied on any GCN-based model since it is model-agnostic. Extensive experiments on three challenging benchmarks prove the effectiveness and generalizability of our approach.

\bibliographystyle{ACM-Reference-Format}
\bibliography{acmart}
\clearpage
\section*{A1. Additional Quantitative Results}
In this section, we provide additional experimental results for further attaining insight into our proposed method, \textit{i.e.,} Adaptive Cross-Form Learning (ACFL). For fairness, all additional experiments are conducted on NTU-RGB+D 120 (X-Sub). Details are as follows.

\noindent\textbf{(1) Effect of Regulatory Factor $\beta$.} As described in Section 3.2, the regulatory factor $\beta$ is adopted to re-adjust the importance weights deciding which source model is the reference one. Here, we would like to investigate the effect of the regulatory factor $\beta$ in ACFL. In particular, we conduct diagnostic experiments under two settings: 1) The On-line ACFL either with the factor or without the factor during training; and 2) The Off-line ACFL either with the factor or without the factor during training. Furthermore, we adopt two representative GCNs for this investigation, i.e., CTR-GCN and Shift-GCN. The experimental results are summarized in Tab.~\ref{tab:abla_study_beta}. 
From the results, we can have two observations: 1) Compared with the model trained via ACFL but without the regulatory factor, adopting the regulatory factor to re-adjust the importance weights of source models will lead to better or comparable results. This shows the effectiveness of the regulatory factor for assessing the quality of source models to some extent. 2) When comparing the On-line ACFL with Off-line ACFL, the performance gains coming from the regulatory factor in the former one are higher than the counterpart in the latter one. We conjecture the possible reason is that the quality of the regulatory factor in the Off-line ACFL is stably high since it represents the accuracy scores produced from pre-trained source models.

\noindent\textbf{(2) Ablation Study of Cross-Modalities.} As a supplementary to the ablation study of cross-form mimicking learning in Section 4.4, we conduct an extra ablation study to investigate whether our proposed ACFL can benefit GCN models under the challenging cross-modality setting. In this work, we consider the modality of heatmap-based RGB image as the one we study. Specifically, two standard GCN models are adopted for investigation, i.e., CTR-GCN and Shift-GCN. The topological architectures of various single-form models are summarized in Tab.~\ref{tab:model_architecture}, which involves the specification of layers and the output size. The corresponding performance results are reported in Tab.~\ref{tab:abla_study_cross_modality}. Not surprisingly, the ACFL significantly improves the heatmap-based models by a margin, i.e., 3.9\% gains for HM-CTR-GCN and 3.6\% for HM-Shift-GCN. This clearly shows the effectiveness of the ACFL for exploiting model potential under the cross-modality setting. Notably, mimicking the representations derived from both $joint$ and $bone$ based skeleton forms can respectively improve the heatmap modality based target models, i.e., HM-CTR-GCN and HM-Shift-GCN, by 3.5\% and 3.3\% accuracy. However, the performance gains start to saturate when mimicking the representations derived from all skeleton forms.   

\begin{table}[ht]
	\centering
	\caption{Additional ablation studies of cross-form mimicking learning. Investigating the effect of the regulatory factor $\beta$ in ACFL.}
	\vspace{-0.em}
	\setlength{\belowcaptionskip}{-0.5cm}%
	\renewcommand\arraystretch{1.2}
	\resizebox{0.9\linewidth}{!}{
		\begin{tabular}{cc|ccc|ccc}
			
			\hline
			\multicolumn{2}{c|}{On-line ACFL}     &  \multicolumn{3}{c|}{CTR-GCN} &  \multicolumn{3}{c}{Shift-GCN}  \\ \hline
			w/o $\beta$   & w $\beta$ &  $joint$ & $bone$ & $both$ &  $joint$ & $bone$ & $both$\\ 
			\hline
			\checkmark  &  & 85.6\% & 87.2\%  &  88.4\% & 84.7\% & 83.6\% &  85.2\% \\
			& \checkmark & 86.4\% & 87.6\%  &  88.7\% & 84.4\% & 84.7\% & 86.2\% \\
			\hline
			\hline
			\multicolumn{2}{c|}{Off-line ACFL}     &  \multicolumn{3}{c|}{CTR-GCN} &  \multicolumn{3}{c}{Shift-GCN}       \\ \hline
			w/o $\beta$   & w $\beta$ &  $joint$ & $bone$ & $both$ &  $joint$ & $bone$ & $both$\\ 
			\hline
			\checkmark  &  & 86.8\% & 88.2\%  & 89.1\%  & 84.9\% & 85.7\%  & 87.0\% \\
			& \checkmark & 87.3\% & 88.4\%  &  89.3\% & 85.1\%  & 85.5\% & 86.9\% \\
			\hline
		\end{tabular}
	}
	\label{tab:abla_study_beta}
\end{table}

\begin{table}[ht]
	\centering
	\caption{Additional ablation studies of cross-form mimicking learning. Investigating the effect of the ACFL under the cross-modality setting.}
	\vspace{-0.em}
	\setlength{\belowcaptionskip}{-0.5cm}%
	\renewcommand\arraystretch{1.2}
	\resizebox{1.0\linewidth}{!}{
		\begin{tabular}{c|ccc|c}
			
			\hline
			\multirow{2}{*}{Target model}       &  \multicolumn{3}{c|}{Source model}   &   \multirow{2}{*}{Acc}      \\ \cline{2-4}
			&  \multicolumn{1}{c|}{J-CTR-GCN}   & \multicolumn{1}{c|}{B-CTR-GCN} & \multicolumn{1}{c|}{BJ-CTR-GCN} & \\ 
			\hline
			\hline
			\multirow{3}{*}{HM-CTR-GCN}   & - & - & -  &  73.4\% \\\cline{2-5}
			& \checkmark & \checkmark  &   & 76.9\%  \\
			& \checkmark &  \checkmark & \checkmark  &  77.3\%  \\
			\hline
			\hline
			\multirow{2}{*}{Target model}       &  \multicolumn{3}{c|}{Source model}   &   \multirow{2}{*}{Acc}      \\ \cline{2-4}
			&  \multicolumn{1}{c|}{J-Shift-GCN}   & \multicolumn{1}{c|}{B-Shift-GCN} & \multicolumn{1}{c|}{BJ-Shift-GCN} & \\
			\hline
			\multirow{3}{*}{HM-Shift-GCN}   & - & - &-   &  72.2\% \\\cline{2-5}
			& \checkmark & \checkmark &   &  75.5\%  \\
			& \checkmark &  \checkmark & \checkmark  & 75.8\% \\
			\hline
		\end{tabular}
	}
	\label{tab:abla_study_cross_modality}
\end{table}

\begin{table*}[ht]
	\centering
	
	\caption{An instantiation of the GCN networks based on various skeleton forms.  HM denotes the heatmap, while BN is the batch normalization. The GAP denotes the global average pooling and FC is the fully-connected layer.}
	\renewcommand\arraystretch{1.3}
	\resizebox{0.99\linewidth}{!}{
		\begin{tabular}{c|c|c|c|c}
			\hline
			Stage      & Heatmap Image HM-GCN              & Joint/Bone J/B-GCN       & Joint \& Bone BJ-GCN     & Output sizes    \\ \hline
			Input & uniform, 64 frames              & uniform, 64 frames          & uniform, 64 frames          &  $\begin{array}{cc} & HM: B\times1\times64\times50\times56 \\ & Joint/Bone: B\times3\times64\times25\times2 \\ & Joint\&Bone: B\times6\times64\times25\times2 \end{array} $  \\ \hline
			Stem & $\begin{array}{cc} & PatchEmbedding, \\ & Tokens: (1,25,2), \\ & embedding\quad dim: 64 \end{array}$ & BN, $3\times25\times2$               & BN, $6\times25\times2$               & $\begin{array}{cc} & HM: B\times64\times64\times25\times2 \\ & Joint/Bone: B\times3\times64\times25\times2 \\ & Joint\&Bone: B\times3\times64\times25\times2 \end{array} $ \\ \hline
			$GCN_1$       & $\left[\begin{array}{ccc}&GCN\_unit, 64& \\&TCN\_unit, 64& \end{array}\right]\times1$    & $\left[\begin{array}{ccc}&GCN\_unit, 64& \\&TCN\_unit, 64& \end{array}\right]\times1$  & $\left[\begin{array}{ccc}&GCN\_unit, 64& \\&TCN\_unit, 64& \end{array}\right]\times1$  & $\begin{array}{cc} & HM: B\times64\times16\times25\times2 \\ & Joint/Bone: B\times64\times32\times25\times2 \\ & Joint\&Bone: B\times64\times32\times25\times2 \end{array} $ \\ \hline  
			$GCN_2$       & $\left[\begin{array}{ccc}&GCN\_unit, 64& \\&TCN\_unit, 64& \end{array}\right]\times3$    & $\left[\begin{array}{ccc}&GCN\_unit, 64& \\&TCN\_unit, 64& \end{array}\right]\times3$  & $\left[\begin{array}{ccc}&GCN\_unit, 64& \\&TCN\_unit, 64& \end{array}\right]\times3$  & $\begin{array}{cc} & HM: B\times64\times16\times25\times2 \\ & Joint/Bone: B\times64\times32\times25\times2 \\ & Joint\&Bone: B\times64\times32\times25\times2 \end{array} $ \\ \hline 
			$GCN_3$       & $\left[\begin{array}{ccc}&GCN\_unit, 128& \\&TCN\_unit, 128& \end{array}\right]\times3$    & $\left[\begin{array}{ccc}&GCN\_unit, 128& \\&TCN\_unit, 128& \end{array}\right]\times3$  & $\left[\begin{array}{ccc}&GCN\_unit, 128& \\&TCN\_unit, 128& \end{array}\right]\times3$  & $\begin{array}{cc} & HM: B\times128\times8\times25\times2 \\ & Joint/Bone: B\times128\times16\times25\times2 \\ & Joint\&Bone: B\times128\times16\times25\times2 \end{array} $ \\ \hline
			$GCN_4$       & $\left[\begin{array}{ccc}&GCN\_unit, 256& \\&TCN\_unit, 256& \end{array}\right]\times3$    & $\left[\begin{array}{ccc}&GCN\_unit, 256& \\&TCN\_unit, 256& \end{array}\right]\times3$  & $\left[\begin{array}{ccc}&GCN\_unit, 256& \\&TCN\_unit, 256& \end{array}\right]\times3$  & $\begin{array}{cc} & HM: B\times256\times4\times25\times2 \\ & Joint/Bone: B\times256\times8\times25\times2 \\ & Joint\&Bone: B\times256\times8\times25\times2 \end{array} $ \\ \hline
			Output       & GAP, FC    & GAP, FC   & GAP, FC   & $\mathbf{\#}$ classes \\ \hline
			
		\end{tabular}
	}
	\label{tab:model_architecture}
\end{table*}

\begin{figure*}[t]
	\centering
	\setlength{\abovecaptionskip}{-0.cm}%
	\includegraphics[width=0.99\linewidth]{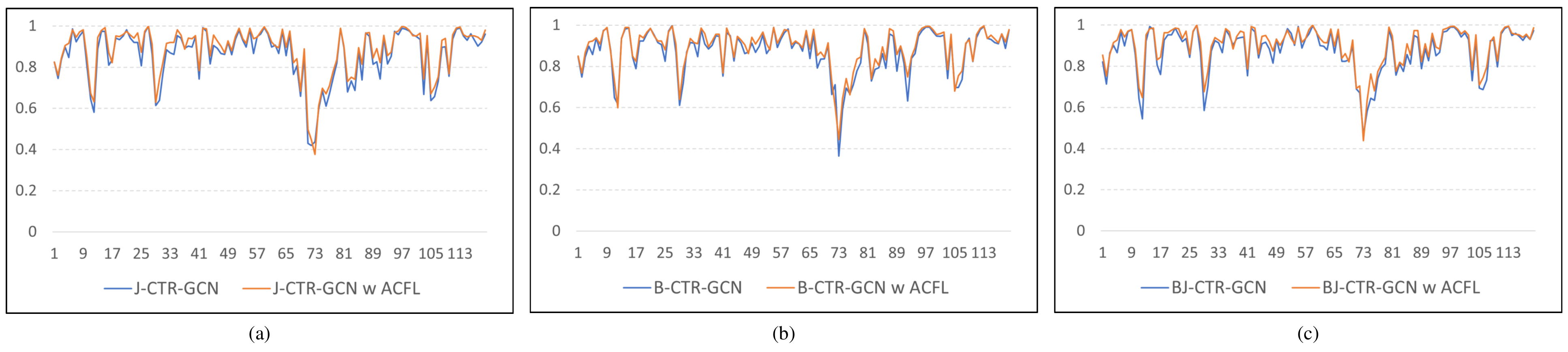}
	\caption{Class-wise accuracy performance evaluated on NTU RGB-D 120 dataset: (a) joint based CTR-GCN models; (b) bone based CTR-GCN models; (c) hybrid form based CTR-GCN models. All GCN models trained via our proposed ACFL outperform baseline GCNs for most of the action classes. All GCNs perform worst at hard action classes that index from 70 to 75, which are hand-related actions.}
	\label{fig:acfl_comp}
\end{figure*}

\begin{figure*}[t]
	\setlength{\abovecaptionskip}{-0.cm}%
	\centering
	\includegraphics[width=0.99\linewidth]{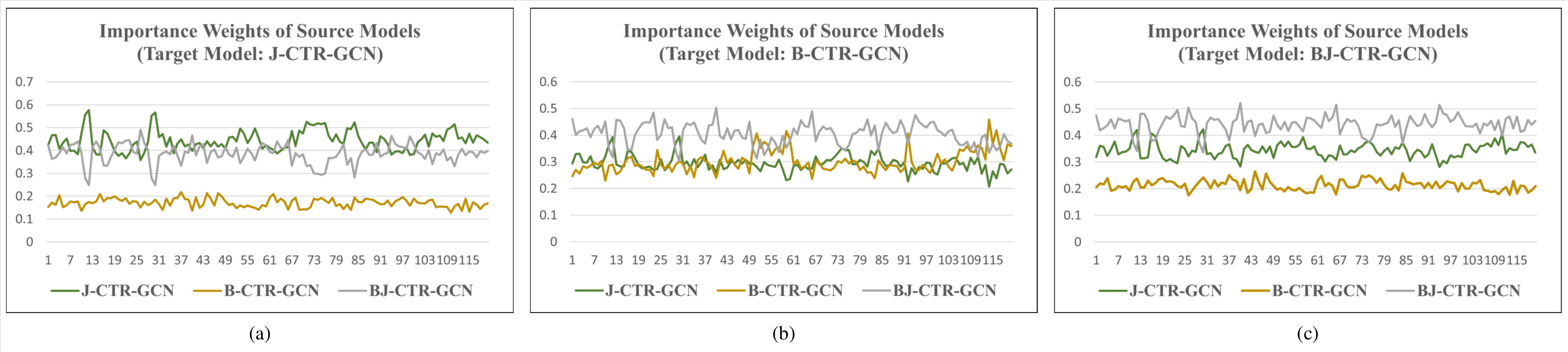}
	\caption{Class-wise importance weights of source models evaluated on NTU RGB-D 120 dataset. Each curve indicates how much each source model contributes to the generation of complementary representation for the specific target model: (a) joint based CTR-GCN models; (b) bone based CTR-GCN models; (c) hybrid form based CTR-GCN models.}
	\label{fig:acfl_comp_atten}
\end{figure*}

\section*{A2. Additional Qualitative Results}
In this section, we conduct a class-wise evaluation and calculate the per-class accuracy. As shown in Fig.~\ref{fig:acfl_comp}, we use the CTR-GCN model for investigation and adopt two settings for comparison, i.e., either with or without ACFL in the training phase. Corresponding accuracy curves spotted in Fig.~\ref{fig:acfl_comp} intuitively indicate the overall effectiveness of our proposed ACFL as well as the challenges that GCN models are suffering from. Specifically, on the one hand, all GCN models trained via the ACFL outperform baseline models for most of the action classes. On the other hand, in line with findings stated in Section 4.4, all GCN models perform worst at hard action classes that index from 70 to 75, where those classes are characterized by hand-related actions with minor changes. To tackle these, other cues like RGB appearance as well as optical flow should be also considered for recognition. Besides, we also plot the curves of importance weights in Fig.~\ref{fig:acfl_comp_atten}, indicating which source model should be considered for each target model in ACFL. From the results, we have the following observations: 1) all source models are required to generate complementary action representation. This indicates that representations derived from different forms of skeletons can complement each other, which is in line with our motivation stated in the above section. 2) Different target models have different learning tendencies to the source models. For example, when J-CTR-GCN is set as the target model, it prefers to mimic the representations from itself and BJ-CTR-GCN, but the importance weight of B-CTR-GCN is relatively small. Similar trends are also observed when the B-CTR-GCN or BJ-CTR-GCN are set as the target model. 
\end{document}